\documentclass[10pt,twocolumn,letterpaper]{article}

\usepackage[pagenumbers]{cvpr} % To force page numbers, e.g. for an arXiv version

\usepackage{graphicx}
\usepackage{amsmath}
\usepackage{amssymb}
\usepackage{booktabs}
\usepackage{multirow}
\usepackage{amsmath}
\usepackage[dvipsnames,table,xcdraw]{xcolor}
\usepackage[pagebackref=false,breaklinks,colorlinks,citecolor=RoyalBlue,bookmarks=false]{hyperref}

\usepackage[capitalize]{cleveref}
\crefname{section}{Sec.}{Secs.}
\Crefname{section}{Section}{Sections}
\Crefname{table}{Table}{Tables}
\crefname{table}{Tab.}{Tabs.}

\def\cvprPaperID{7751} % *** Enter the CVPR Paper ID here
\def\confName{CVPR}
\def\confYear{2023}

\begin{document}

\title{HumanRecon: Neural Reconstruction of Dynamic Human Using \\ Geometric Cues and Physical Priors}

\author{Junhui Yin\thanks{Junhui Yin, Zhanyu Ma, Jun Guo are with Beijing University of
Posts and Telecommunications, China.}
\and
Wei Yin\thanks{Wei Yin is with the DJI Technology, China.}
\and
Hao Chen\thanks{Hao Chen is with Zhejiang University, China.}
\and
Xuqian Ren\thanks{Xuqian Ren is with Tampere Universities, Finland.}
\and
Zhanyu Ma\footnotemark[1]
\and
Jun Guo\footnotemark[1]
\and
Yifan Liu\thanks{Yifan Liu is with University of Adelaide, Australia.}}
\maketitle

%%%%%%%%% ABSTRACT
\begin{abstract}
Recent 
methods for dynamic human reconstruction have attained 
promising 
reconstruction results.
%however their performance relies 
Most of these methods 
rely only on RGB color supervision without considering explicit geometric constraints. This leads to existing human reconstruction techniques being more prone to overfitting to color and 
causes 
geometrically inherent ambiguities, especially in the sparse multi-view setup.

Motivated by recent advances in the field of monocular 
geometry 
prediction, we consider the geometric constraints of estimated depth and normals 
in
the learning of neural implicit representation
for dynamic human reconstruction. As a geometric regularization, this provides reliable yet explicit 
supervision 
information, 
and improves reconstruction quality. We also exploit several % well-grounded 
beneficial 
physical priors, such as adding noise into view direction and maximizing the density on the human surface. 
These priors ensure 
the color rendered along rays
to be 
robust to view direction and reduce the inherent ambiguities of density estimated along rays. Experimental results demonstrate that depth and normal cues, predicted by human-specific monocular estimators, can provide effective supervision signals and render more accurate images. Finally, we also show that the proposed physical priors significantly reduce overfitting and improve the overall quality of novel view synthesis. Our code is available at:~\href{https://github.com/PRIS-CV/HumanRecon}{https://github.com/PRIS-CV/HumanRecon}.
\end{abstract}

\begin{figure}[t]
\begin{center}
\includegraphics[width=0.48\textwidth]{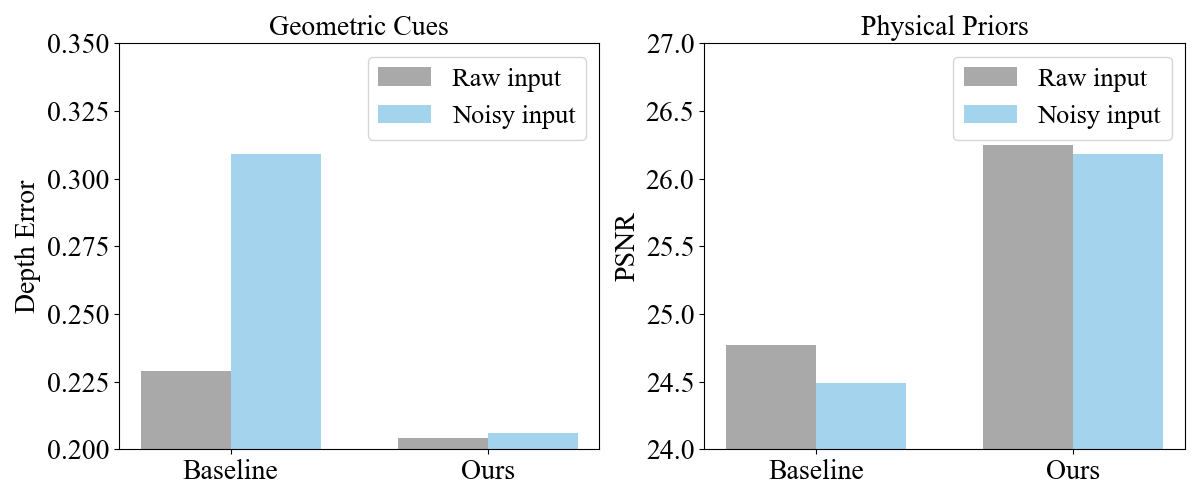}
\end{center}
\vspace{-7mm}
   \caption{With geometric cues, our method is able to greatly improve geometric smoothness (Depth Error$\downarrow$). In addition, the integration of physical priors into the rendering process also greatly improves the image rendering quality (PSNR$\uparrow$). In particular, when the input image carries noise, our method still achieves good results and does not suffer from significant performance degradation as the baseline method.}
\label{intro_fig}
\end{figure}

\section{Introduction}
\label{sec:intro}
3D reconstruction and animation of clothed human avatars is a fundamental problem in computer vision research with various applications in robotics, graphics, augmented reality, virtual reality, and human digitization. In all these applications, the human reconstruction models are expected to achieve high-quality human geometry and appearance performance. Thus, they require various hardware, such as 4D scanners, depth sensors, or RGB cameras. Among them, the captured images and videos by RGB cameras are the most readily available and used. However, they provide the insufficient supervision information, making this setup the most challenging for high-fidelity digital human reconstruction.

Traditional works in human body modeling use explicit mesh to represent the human geometry and store the appearance in 2D texture maps, but they require dense camera arrays~\cite{DeepBlending2018,joo2018total} and controlled lighting conditions~\cite{collet2015high,guo2019relightables}. These devices are extremely bulky and expensive for daily use. In contrast to these works, PIFu~\cite{saito2019pifu}, StereoPIFu~\cite{hong2021stereopifu} and PIFuHD~\cite{saito2020pifuhd}
propose to regress neural implicit function using pixel-level image features and are able to reconstruct high-resolution clothed human results.
ARCH~\cite{huang2020arch} extends a series of PIFu methods to regress animatable clothed human avatars in a canonical pose from monocular images. These methods adopt neural implicit function to represent the human’s implicit geometry and achieve impressive results, but fail to reconstruct dynamic clothed humans from a sparse set of multi-view videos. In addition, they also require the corresponding ground-truth geometry of color images to train the model, which is expensive to acquire and limits their generalization.

\begin{figure*}[t]
\begin{center}
\includegraphics[width=0.9\textwidth]{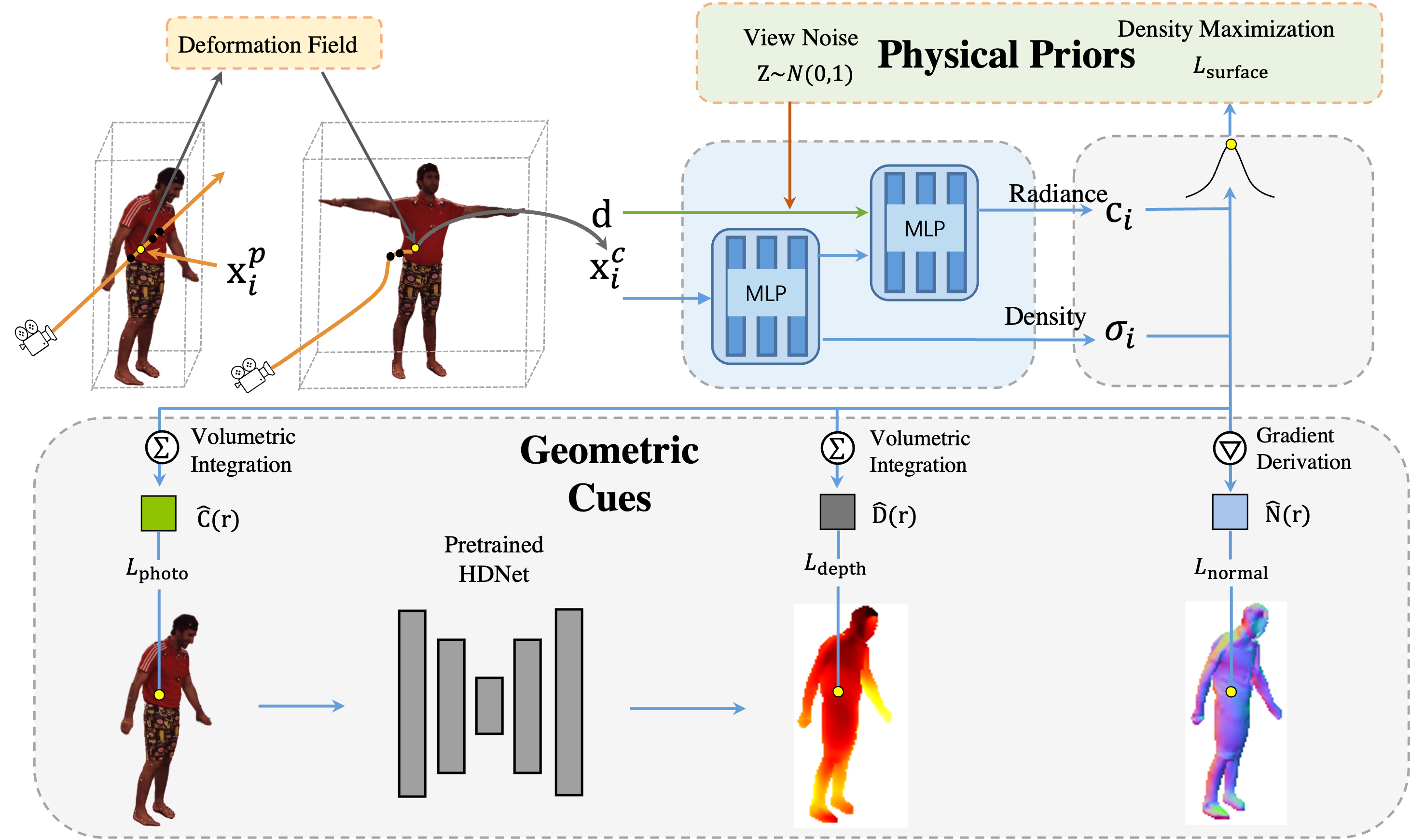}
\end{center}
\vspace{-7mm}
   \caption{The overview of our method. In this work we use the geometric cues of estimated depth and normals, and physical priors to guide the optimization of neural implicit human models. More specifically, for a batch of rays, we render  their predicted color and expected termination depth, and calculate the gradient of volume density as surface normal.
   Given the geometry properties predicted by pretrained network~\cite{jafarian2021learning}, we supervise the rendered geometry results by comparing them with the corresponding estimation. In addition, we apply carefully-designed training strategies (i.e. view noise and density maximization) from physical priors to ensure the rendered color robust for view direction and reduce the inherent ambiguities of density estimated along rays.
   }
\label{overview}
\end{figure*}

Recently, neural radiance fields (NeRF)~\cite{mildenhall2021nerf}, as an effective scene representation, achieve impressive novel view synthesis results with a differentiable renderer. Many methods~\cite{peng2021animatable,noguchi2021neural} use NeRF as an implicit representation to remove the need for ground-truth geometry and reconstruct humans from sparse multi-view videos with only image supervision. 
Animatable NeRF~\cite{peng2021animatable} uses a parametric human body model as a strong geometry prior to canonical NeRF model, and achieves impressive visual fidelity on novel view and pose synthesis results. NARF~\cite{noguchi2021neural} proposes neural deformable fields to model the dynamic human and represent pose-controllable changes of articulated objects.
% However, there are two fundamental drawbacks of these existing approaches. 
However, these existing methods usually suffer from two problems: (1) the NeRF-based reconstruction results tend to be noisy due to the lack of proper geometric regularization on radiance fields. This is more common in sparse-view dynamic human reconstructions and often results in erroneous artifacts in the form of color blocks. 
(2) As existing approaches train their canonicalization networks on inputs in observation space, they are more prone to overfitting to color. The rendering quality depends heavily on the color quality of the observed image, leading to a severely performance degradation when the raw image appears significantly noisy, as shown in the Fig.~\ref{overview}.

Motivated by recent advances in the field of human-specific monocular geometry prediction, we leverage the estimated geometric cues as a regularization technique in dynamic human reconstruction.
The predicted pseudo ground-truths provide globally consistent and smooth geometric constraints as a complement to the photometric consistency RGB loss during optimization. 
% together with the photometric consistency RGB loss. 
We observe that the geometry-based regularization technique leads to significant improvement in geometry reconstruction quality as shown in Fig.~\ref{intro_fig}. This is due to the fact that depth and normal constraints can mitigate the geometry ambiguity and achieve smoother reconstructions with more fine details, required by image-based view synthesis.

Apart from incorporating these geometric cues, we investigate the strong physics background of neural radiation rendering and elaborately select two physical priors. Based on these physical priors, we construct additional optimization objective functions to guide the neural reconstruction process. The key idea is that good rendering results should be highly robust to view direction and the estimated density along rays should be the maximum value at the ray-surface points\footnote{Intersection points of the camera ray and the human surface.}. We leverage this idea as proxy guidance to help our model to learn view-invariant representation and reduce the inherent ambiguities of density estimated along rays.
By integrating geometric cues and physical priors into human reconstruction process, we can successfully constrain both geometry and appearance to boost the fidelity of neural radiance fields even in highly sparse multi-view setting.

In summary, we make the following contributions:

\begin{itemize}
\itemsep 0cm
\item

We propose the HumanRecon approach, which leverages geometric cues and physical priors to improve geometry reconstruction and photorealistic rendering.

\item
To enforce surface constraints on the learned geometry, we use depth and surface normals, predicted by a human-specific geometry estimator, as additional supervision signals during volume rendering together with the RGB image reconstruction loss.

\item 
We investigate the strong physics background of neural radiation
rendering and elaborately select several physical prior as proxy guidance, which help our model to alleviate the overfitting on image color and reduce the inherent ambiguities of density estimated along rays.

\item
We conduct extensive experiments on several challenging datasets, and show the effectiveness and interpretability of our method with quantitative and qualitative results.

\end{itemize}

\section{Related Work}
\label{sec:Related Work}

\textbf{3D human reconstruction}. The 3D shape of a human body is typically represented by parametric mesh models such as SCAPE~\cite{anguelov2005scape} or SMPL~\cite{loper2015smpl}.
SMPL recovers the human mesh 
% given the 3D scans and 
deforms the template mesh with the linear blend skinning (LBS) algorithm. The template-based 3D human model~\cite{zhang2021pymaf,kolotouros2019learning} is easy to control huamn pose, but difficulty to model large garment deformations. Unlike mesh-based method, 
%deep implicit functions can represent 3D shapes with arbitrary topology changes and reconstruct
%high-resolution clothed human results. 
PIFu~\cite{saito2019pifu} learns an implicit surface function using pixel-level image features and generalizes well to clothing-type variation.
Geo-PIFu~\cite{he2020geo} adds a coarse human shape proxy, and Dong et al.~\cite{xu2022geo} uses a depth denoising process to regularize global shape. 
% and learn robust geometry information.
Recent methods~\cite{huang2020arch,he2021arch++,bhatnagar2020combining,bhatnagar2020loopreg,zheng2021pamir} use both mesh-based statistical models and deep implicit functions and achieve impressive performance. 
For example, ARCH~\cite{huang2020arch} and ARCH++~\cite{he2021arch++} combine implicit function learning with the deformation field to achieve the animatable reconstruction of clothed humans.
Such methods often require high-quality 3D geometry or dense multi-view images with depth for training, while our goal is to train model on sparse-view image-level RGB supervision alone.

\textbf{Neural radiance fields for a dynamic human body}. Our work builds on a new 3D scene representation form of neural radiance fields (NeRF)~\cite{mildenhall2021nerf}.
NeRF and its extensions~\cite{barron2021mip,hedman2021baking,wang2021ibrnet,fridovich2022plenoxels} are capable of rendering novel views of static scenes with high quality due to their powerful expressiveness.
Recent works~\cite{noguchi2021neural,peng2021neural,peng2021animatable,peng2022animatable,liu2021neural} add a deformation field on the original NeRF to enable it to model dynamic human body. Some works~\cite{peng2021animatable,peng2022animatable} use the human pose and skinning weights form SMPL as prior knowledge to facilitate the learning of deformation field. NeuralBody~\cite{peng2021neural} use the same
set of latent codes anchored at deformable mesh to encode the pose-deformation observations. Further, NARF~\cite{noguchi2021neural} resorts to human skeleton to learn pose-dependent changes from images with pose annotations. These approaches mainly focus on how to better learn a deformation field and animate human pose, while they do not fully utilize geometric cues or physical priors to enhance the rendering capabilities of neural human reconstruction model.

\textbf{Incorporating priors into neural scene representations}.
Recent studies employ depth priors~\cite{azinovic2022neural}, sparse point clouds~\cite{roessle2022dense}, semantic labels~\cite{guo2022neural}, and Manhattan-world Assumption~\cite{guo2022neural,coughlan1999manhattan} for novel view synthesis on neural scene representations. These methods are beneficial for static indoor 3D scene such as walls and floors, but difficult to handle dynamic human body. Motivated by recent advances in the field of human-specific geometry prediction, we use the estimated human depth and normals as geometric cues for dynamic human reconstruction. Based on the observation of neural rendering process, we also integrate physical priors into human reconstruction model and further demonstrate their effectiveness.

\section{Method}
\label{sec:Met}

Our goal is to reconstruct a clothed human model that can be used to render free-view human videos while utilizing geometric cues and physical priors to guide the rendering process. To this end, we start by briefly reviewing \textbf{Ani}matable \textbf{Ne}ural \textbf{R}adiance \textbf{F}ields (Ani-NeRF)~\cite{peng2021animatable} as our baseline model in Section 3.1. Next, we introduce the predicted geometric cues we investigate, and formulate their loss functions to integrate pseudo ground-truth labels from a pretrained network into our baseline model in Section 3.2. Finally, to alleviate the overfitting to image colors and reduce density ambiguity along rays, we leverage several beneficial physical priors into a prior loss committee as the proxy guidance for neural human reconstruction in section 3.3. An overview of our proposed method is presented in Fig.~\ref{overview}.

\subsection{Revisiting Animatable Neural Radiance Fields}
% \label{sec:exper}
Ani-NeRF firstly represents the dynamic human body in the video with the posed space and the canonical space. The posed space is defined for each video frame, and the deformation field $T$ maps each posed space point $\textbf{x}$ to the canonical space. Then, Ani-NeRF takes spatial location $\textbf{x}$ and viewing direction $\textbf{d}$ as input to a shared canonical human network and outputs a view-invariant volume density $\sigma$ and view-dependent emitted radiance $\textbf{c}$.

$\textbf{Deformation field}$. To be more specific about the deformation field $T$, we first define the posed space for each frame video and the canonical space shared for all the frames. For a pose-space point $\textbf{x}^{p}$ at frame $i$, we have rigid transformations of body joints, ${G_{i}^{k}} \in SE(3)$, which are from the fitted SMPL model~\cite{loper2015smpl}. Based on this transformation, the spatial point $\textbf{x}^{p}$ can be mapped to the canonical space by $\textbf{x}^{c}=\left ( \sum_{k=1}^{K} w_{i}^{k}(x)G_{i}^{k} \right )^{-1} \textbf{x}^{p}$. $\textbf{w}_{i}(x)=\{w_{i}^{k}(x)\}_{k=1}^{K}$ is the blend weight function of $k$-part. 
Note that, a neural blend weight $\textbf{w}^{\textbf{can}}$ at the canonical space should also be additionally learned to animate the human NeRF. 
More details on this part can be found in~\cite{peng2021animatable}. 

\textbf{Canonical human network}. Canonical human network~\cite{mildenhall2021nerf} represents a static scene as a continuous volumetric representation learned by a neural network. Specifically, given the density network $F_{\sigma}$, the density model of human body at frame $i$ can be defined as 
\begin{equation}
  \left ( \sigma _{i},\textbf{c}_{i} \right ) =F_{\sigma } \left (\gamma_{\textbf{x}}( T_{i}\left ( \textbf{x} \right )   \right ), \gamma_{\textbf{d}}(\textbf{x})),
\end{equation}
where the positional encoding $\gamma$~\cite{mildenhall2021nerf} transforms the sampled coordinate and view direction into their Fourier features that facilitate learning of high-frequency details.
Given a pixel, a ray $\textbf{r}(\tau)=\textbf{o}+\tau\textbf{d}$ is cast from the camera origin \textbf{o} and through the pixel along direction \textbf{d}. Along this ray, we sample $J$ discrete points $\{\textbf{x}_j=\textbf{r}(\tau_j)\}_{j=1}^{J}$ and estimate their densities and color $\{\sigma_{j},\textbf{c}_j\}_{j=1}^{J}$. The predicted color of this pixel can be rendered by a numerical quadrature approximation~\cite{max1995optical}:
\begin{equation}\label{volumetic render}
\hat{\textbf{C}}(\textbf{r})=\sum_{j=1}^{J}\Gamma_j(1-\exp(-\sigma_k(\tau_{j+1}-\tau_{j})))\textbf{c}_k, 
\end{equation}
\quad \quad \quad \quad \quad with $\Gamma_k=\exp(-\sum_{j^{'}<j}(\tau_{j^{'}+1}-\tau_{j^{'}} ) )$ \\

\noindent where $\{(\tau_j)\}_{j=1}^{J}$ are discrete sampling points on the ray \textbf{r} and $\Gamma_j$ can be interpreted as an accumulated transmittance along the ray. Finally, Ani-NeRF is trained to minimize the following photometric loss and blend weight consistency loss.
\begin{equation}
L_{\text{photo}}=\sum_{\textbf{r}\in \mathcal{R}}\|\hat{\textbf{C}}(\textbf{r})-\textbf{C}(\textbf{r})\|_{2}, 
\end{equation}
and 
\begin{equation}
L_{\text{weight}}=\sum_{\textbf{x}\in \mathcal{X}_i}\|\textbf{w}_i(\textbf{x})-\textbf{w}^{\text{can}}(T_{i}(\textbf{x}))  \|_{1}, 
\end{equation}
\noindent where $\mathcal{R}$ represents a batch of training rays passing through image pixels and $\mathcal{X}_{i}$ is the set of 3D points sampled within the 3D human bounding box at frame $i$.
$L_{\text{photo}}$ enforces the rendered pixel color  $\hat{\textbf{C}}(\textbf{r})$ to be consistent with the observed 
pixel color $\textbf{C}(\textbf{r})$, while $L_{\text{weight}}$ minimizing the difference between blend weight fields.
$L_{\text{base}}=L_{\text{photo}}+L_{\text{weight}}$ is the overall optimization objective of Ani-NeRF. For more details, we refer readers to~\cite{peng2021animatable}. 

\subsection{Pseudo-supervision from Geometric Cues}

As a powerful representation, Ani-NeRF produces impressive photorealistic renderings of unseen viewpoints. Yet, this approach relies only on RGB color supervision without considering explicit geometric constraints, leading to es geometrically inherent ambiguities in sparse-view videos. To overcome this limitation, we unify Ani-NeRF's powerful implicit rendering capabilities with explicit 3D reconstruction methods by using explicit geometry supervision.

\textbf{Predicted depth cues}. One common geometric cues is the depth map, which contains geometric structures of shapes in 3D space and can be easily obtained with an off-the-shelf depth estimator. Specifically, we use a pretrained human depth estimator~\cite{jafarian2021learning} to predict a depth map $D$, which is faithful to the input real image. Our goal is to use the predicted depth cues as a regularization technique thereby improving existing neural implicit methods. A key observation is that the volume rendering scheme can generate depth maps for each image in a similar manner to rendering RGB pixels. Formulately, the expected termination depth of each ray \textbf{r} can be produced by modifying volumetric rendering equation (i.e.  Eq.~\ref{volumetic render}) as 
\begin{equation}
\hat{\textbf{D}}(\textbf{r})=\sum_{j=1}^{J}\Gamma_j(1-\exp(-\sigma_k(\tau_{j+1}-\tau_{j})))\tau_k.
~\label{eq:depth estimation}
\end{equation}
where $\{(\tau_j)\}_{j=1}^{J}$ are discrete sampling points on the ray \textbf{r}.

A straightforward strategy is to use depth-based pseudo-label to directly supervise the learning of this depth estimated by Ani-NeRF via the squared distance. $\hat{\textbf{D}}$ and $\textbf{D}$ are both absolute depth estimates from their own networks, and therefore hinders depth supervision in general scenes. To alleviate the influence, we consider depth information as relative cues and enforce consistency between our rendered expected depth and the predicted depth by
\begin{equation}
L_{\text{depth}}=\sum_{\textbf{r}\in \mathcal{R}} \|\frac{\hat{\textbf{D}}(\textbf{r})}{\text{Mean}_{\textbf{r}\in \mathcal{R}}{(\hat{\textbf{D}}(\textbf{r}))}}-\frac{\textbf{D}(\textbf{r})}{\text{Mean}_{\textbf{r}\in \mathcal{R}}{(\textbf{D}(\textbf{r}))}}\|_{2},
\end{equation}
\noindent where $\mathcal{R}$ represents a batch of training rays \textbf{r} passing through image pixels and two depth distributions are scaled with the mean value calculated from themselves.

\textbf{Predicted normal cues}. The surface normal is another geometric cues we use. Unlike depth cues that provide globally structure information and reduce shape ambiguity, normal cues are local and capture smooth geometries with fine details. Similar to the depth cues, we use the same model to predict normal maps \textbf{N} for each RGB image,  which provide constraint on the normals calculated by the gradient of the volume density $\sigma$ with respect to 3D position \textbf{x}~\cite{boss2021nerd}:
\begin{equation}
\hat{\textbf{N}}(\textbf{x})=-\frac{\nabla \sigma(\textbf{x})}{\|\nabla \sigma(\textbf{x})\|}.
\end{equation}

To regularize the consistency of the computed density gradient normals $\hat{\textbf{N}}$ with pseudo ground truth normals $\textbf{N}$ from a pretrained network, we enforce a cosine embedding loss on them:
\begin{equation}
L_{\text{normal}}=1-\cos(\hat{\textbf{N}}(\textbf{x}), \textbf{N}(\textbf{x})).
\end{equation}
Normal cues are local and capture smooth geometric feature, while depth cues provide relative information with strong global constraints. We hence expect that two geometric constraints are complementary and enhance each other.

\begin{table*}[htbp]
  \centering
    \scalebox{0.85}{
    \begin{tabular}{crcccccccccc}
    \toprule
    \multirow{2}[3]{*}{Baseline} & \multicolumn{2}{c}{Geo. Cues} &       & \multicolumn{2}{c}{ Phy. Priors} &       & \multicolumn{3}{c}{Image} &       & \multicolumn{1}{l}{Geometry} \\
\cmidrule{2-3}\cmidrule{5-6}\cmidrule{8-10}\cmidrule{12-12}          & \multicolumn{1}{c}{Depth} & Normal &       & Vnoise & Dmax  &       & \multicolumn{1}{l}{MSE$\downarrow$} & \multicolumn{1}{l}{PSNR$\uparrow$} & \multicolumn{1}{l}{SSIM$\uparrow$} &       & \multicolumn{1}{l}{Dep. Err.$\downarrow$} \\
     \midrule
    $\surd$    &       &       &       &       &       &       & 0.00354 & 24.77 & 0.907 &       & 0.229 \\
    $\surd$    & $\surd$ &       &       &       &       &       & 0.00337 & 24.94 & 0.907 &       & 0.205 \\
    $\surd$     &       & $\surd$     &       &       &       &       & 0.00340 & 24.94 & 0.909 &       & 0.229 \\
    $\surd$     & $\surd$ & $\surd$   &       &       &       &       & 0.00326 & 25.05 & 0.910  &       & \textbf{0.203} \\
    \midrule
    $\surd$     &       &       &       & $\surd$     &       &       & 0.00249 & 26.17 & 0.909 &       & 0.232 \\
    $\surd$     &       &       &       &       & $\surd$     &       & 0.00339 & 24.99 & 0.912 &       & 0.228 \\
    $\surd$     &       &       &       & $\surd$    & $\surd$     &       & \underline{0.00247} & \underline{26.19} & \underline{0.914}  &       & 0.232 \\
    \midrule
    $\surd$     & $\surd$     & $\surd$     &       & $\surd$     & $\surd$    &       & \textbf{0.00244} & \textbf{26.25} & \textbf{0.916} &       & \underline{0.204}  \\
    \bottomrule
    \end{tabular}}
\caption{Ablation of geometric cues and physical priors on the subject ``S9" of the H36M dataset. Both geometric cues and physical priors significantly improve human reconstruction quality in terms of all evaluation metrics. With geometry cues, our approach can better reduce geometrically inherent ambiguities and represent the underlying geometry as indicated by the lower depth error. With physical priors, finer details can be captured and the rendering results become better. Using both leads to the best performance. ``Dmax" represents density maximization. Due to space limitations, we present the qualitative and quantitative results of other subjects on novel view synthesis in supplementary.}
\label{tab:geometric and physical}
\end{table*}%

\subsection{Physical Priors} 

NeRF-based 3D human reconstruction is the task of using a set of input images and their camera poses to reconstruct a dynamic human representation capable of rendering unseen views. 
Without sufficient multi-view supervision (i.e. sparse-view video), Ani-NeRF is prone to overfitting on existing data and struggles to accurately learn the density distributions along rays. 
In this section, we present several beneficial physical priors and leverage them into a prior loss committee to build a view-consistent representation and reduce the inherent ambiguities of density. 

\textbf{Perturbation on view direction}. To alleviate the overfitting on multi-view videos, we augment the raw input data to provide necessary constraints on volume rendering. Specifically, we treat available RGB images with view direction  as original data and consider these images with their perturbational view directions as the augmentation data. To generate perturbational view directions (i.e. noisy views), we use Gaussian distribution (e.g. $N(0, 1)$) as perturbation and add it into the raw view direction.
Through this way, we propagate the image information under the available views to their surrounding (noisy) views to better utilize them, thus enabling to build a view-invariant representation. This works
well for clean and noise-free images, which standard NeRF takes as input. Further, our view perturbation strategy can be directly applied on noisy raw input images and is highly robust to the zero-mean distribution of raw noise, which is shown in Tab.~\ref{tab:perturbation on view}.

\textbf{Maximizing density at ray-surface points}. The densities predicted by canonical network describe the geometry of scene and are used as weights in the weighted sum of color values of the sampled points along rays.
For each ray that passes through an image pixel, we expect those colors on the ray-surface points to have the largest contribution for rendered color. Following this intuition, our model should take the maximum density values at these points.
Given depth $\hat{\text{D}}(\textbf{r})$ found via Eq.(\ref{eq:depth estimation}), we can further determine the ray-surface points by
\begin{equation}
\textbf{s}(\textbf{r})=\textbf{o}+\hat{\text{D}}(\textbf{r})\textbf{d},  \textbf{r}\in \mathcal{R},
\end{equation}
\noindent where $\mathcal{R}$ represents a batch of training rays \textbf{r} passing through image pixels. Then, we achieve our goal of density maximization on boundary surface by the following loss:

\begin{equation}
L_{\text{surface}}(\textbf{r})=(1-\sigma (\textbf{s}(\textbf{r}) ))^2, \textbf{r}\in \mathcal{R}.
\end{equation}
where $\sigma (\textbf{s}(\textbf{r})$ is density values predicted by the network $F_{\sigma}$ at ray-surface points \textbf{s}(\textbf{r}).
In this way, we can encourage network model to prevent the formation of spurious and invisible surfaces in the scene.

The overall loss function is
\begin{equation}
L_{\text{overall}}=L_{\text{base}}+\lambda_{\text{depth}}L_{\text{depth}}+\lambda_{\text{normal}}L_{\text{normal}}+\lambda_{\text{surface}}L_{\text{surface}},
\end{equation}
where $\lambda_{\text{depth}}$, $\lambda_{\text{normal}}$, and $\lambda_{\text{surface}}$ are weight parameters.

\begin{table}[t]
  \centering
  \scalebox{0.75}{
    \begin{tabular}{llccc}
    \toprule
    Method &    Mode        & MSE$\downarrow $  & PSNR$\uparrow$   & SSIM$\uparrow$  \\
    \midrule
    Baseline &    N/A       & 0.00401 & 24.90  & 0.908 \\
    \midrule
    \multirow{2}[2]{*}{$\gamma(\textbf{d})+N(0.2,0.5)$} & Train \& Test &        0.00257 & 26.24  & 0.915 \\
          & Train  &      \textbf{ 0.00243} & \textbf{26.35 } & \textbf{0.916} \\
    \multirow{2}[2]{*}{$\gamma(\textbf{d})+N(0.0,0.5)$} & Train \& Test &      0.00263 & 26.02    & 0.908 \\
          & Train  &        0.00261 & 26.06    & 0.913 \\
    \midrule
    \multirow{2}[2]{*}{$\textbf{d}+N(1.0,0.1)$} & Train \& Test &       0.00255 & 26.13  & 0.912 \\
     & Train  &       0.00251 & 26.15  & 0.912 \\
    \multirow{2}[2]{*}{$\textbf{d}+N(1.0,0.5)$} & Train \& Test &        0.00256 & 26.12  & 0.911 \\
     & Train  &       0.00250 & 26.17  & 0.912 \\
    \midrule
    $\textbf{d}_{\text{ray}}+N(0.0,0.5)$ & Train \& Test &        0.01500 & 18.55  & 0.715 \\
    \bottomrule
    \end{tabular}}
  \caption{Results of models trained with different view direction perturbation strategies on the subject ``S9" of H36M dataset. ``Train \& Test" implies that perturbation strategy is activated during the test phase, while ``Train" implies that it is deactivated. \textbf{d} and $\gamma(\textbf{d})$ mean adding perturbations before or after the position embedding $\gamma$, respectively;  $\textbf{d}_{\text{ray}}$ represents the direction of ray; $N(\mu ,\sigma)$ implies Gaussian noise with mean $\mu$ and standard deviation $\sigma$.}
  \label{tab:perturbation on view}
\end{table}

\begin{figure}[t]
\begin{center}
\includegraphics[width=0.41\textwidth]{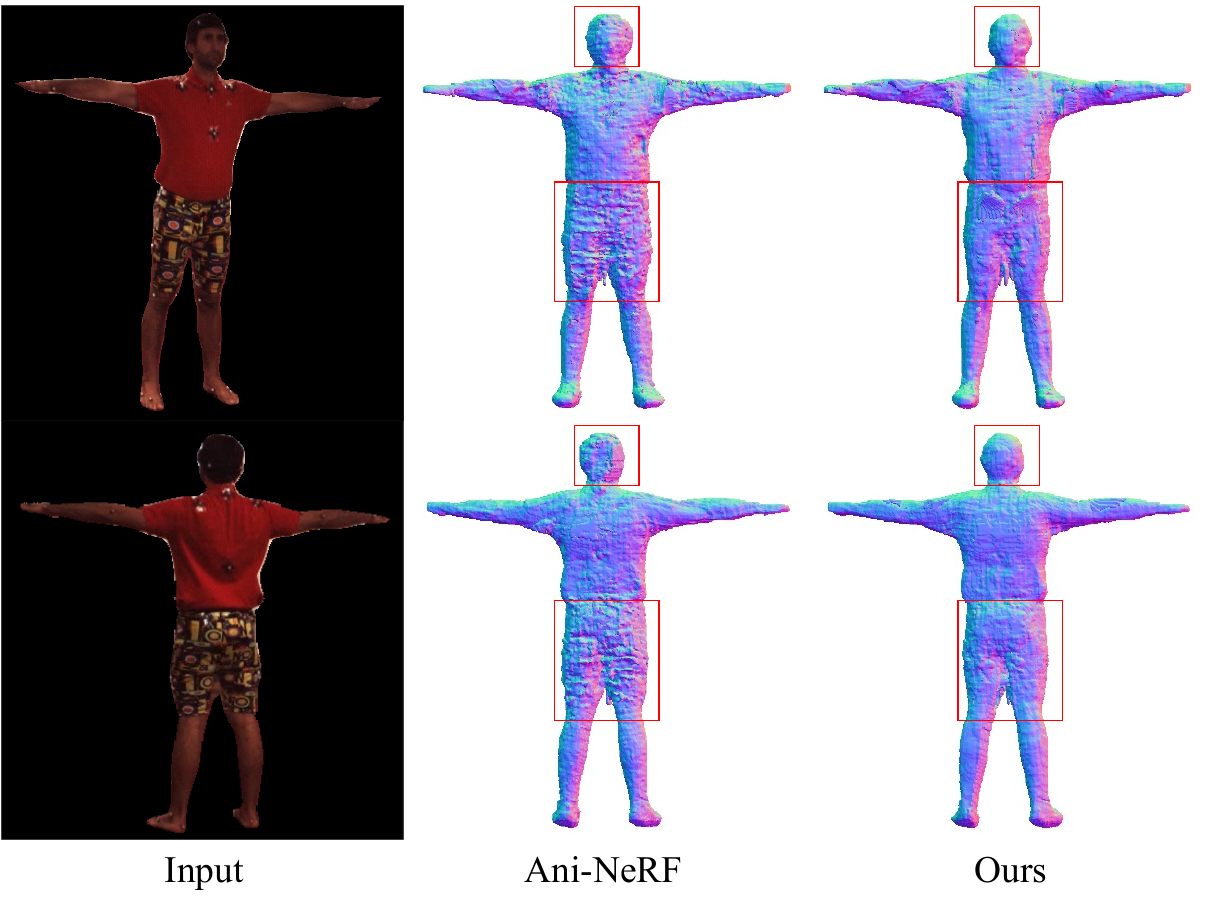}
\end{center}
\vspace{-7mm}
   \caption{Geometry reconstruction results on subject ``S9" of H36M dataset. Compared with Ani-NeRF, our approach reconstructs smoother geometry results with high-quality. Zoom in for details.}
\label{Geometry}
\end{figure}

% \begin{figure}[htb]
% \begin{center}
% \includegraphics[width=0.5\textwidth]{trainnum.pdf}
% \end{center}
% \vspace{-7mm}
%   \caption{}
% \label{intro_fig}
% \end{figure}

\begin{table}[t]
  \centering
        \scalebox{0.7}{
    \begin{tabular}{lccccc}
    \toprule
    \multirow{2}[4]{*}{Method} & \multicolumn{3}{c}{Image} &       & Geometry \\
\cmidrule{2-4}\cmidrule{6-6}          & MSE$\downarrow $   & PSNR$\uparrow $  & SSIM$\uparrow $  &       & Dep. Err.$\downarrow $ \\
    \midrule
    Baseline & 0.00354 & 24.77 & 0.907 &       & 0.229 \\
    \midrule
    Horizontal flip  & 0.00373 & 24.64 & 0.907 &       & 0.233 \\
    Center Cropping & 0.00348 & 24.92 & 0.909 &       & 0.228 \\
    Raw rgb+$N(0, 0.05)$  & 0.00380 & 24.49 & 0.904 &       & 0.234 \\
    \multicolumn{1}{l}{$\textbf{o}_\text{ray} + N(0, 0.5)$}    & 0.01380 & 19.27 & 0.678 &       & N/A \\
     $\textbf{d}_{\text{ray}}+N(0,0.5)$    & 0.01500 & 18.55 & 0.715 &       & N/A \\
     \midrule
HumanRecon & \textbf{0.00244} & \textbf{26.25} & \textbf{0.916} &  &   \textbf{0.204}     \\
    \bottomrule
    \end{tabular}}
  \caption{Comparison between our HumanRecon and other traditional image-based data augmentation methods (e.g. horizontal flips and center cropping). We also show some perturbations on the input data as naive data augmentation, which includes add noise on raw input image (Raw rgb + $N(0,0.05)$, camera origin ($\textbf{o}_\text{ray}$ + $N(0, 0.5)$)), ray direction ($\textbf{d}_{\text{ray}}$ + $N(0,0.5)$). N/A means that the model does not learn any geometric shape}.
  \label{data augmentation}%
\end{table}%
\section{Experiments}
\label{sec:exper}

\begin{table*}[htbp]
  \centering
 \scalebox{0.7}{
    \begin{tabular}{lcrcccccrllccccc}
    \toprule
    \multicolumn{8}{c}{Overfitting}                               &       & \multicolumn{7}{c}{Noisy Input} \\
\cmidrule{1-8}\cmidrule{10-16}    \multirow{2}[4]{*}{Method} & \multirow{2}[4]{*}{N.V} &       & \multicolumn{3}{c}{Image} &       & \multicolumn{1}{l}{Geometry} &       & \multirow{2}[4]{*}{Setting} & \multirow{2}[4]{*}{ Method} & \multicolumn{3}{c}{Image} &       & \multicolumn{1}{l}{Geometry} \\
\cmidrule{4-6}\cmidrule{8-8}\cmidrule{12-14}\cmidrule{16-16}          &       &       & \multicolumn{1}{l}{MSE$\downarrow$} & \multicolumn{1}{l}{PSNR$\uparrow$} & \multicolumn{1}{l}{SSIM$\uparrow$} &       & \multicolumn{1}{l}{Dep. Err.$\downarrow$} &       &       &       & MSE   & PSNR  & SSIM  &       & Dep. Err. \\
\cmidrule{1-8}\cmidrule{10-14}    \multirow{3}[2]{*}{Baseline} & 1     &       & 0.00449 & 23.93 & 0.873 &       & 0.731 &       & \multirow{3}[2]{*}{Raw Input} & Baseline & 0.00354 & 24.77 & 0.907 &       & 0.229 \\
          & 2     &       & \textbf{0.00308} & \textbf{25.35} & \textbf{0.904} &       & 0.456 &       &       & +Phy. Priors & 0.00247 & 26.19 & 0.910  &       & 0.232 \\
          & 3     &       & 0.00354 & 24.77 & 0.907 &       & \textbf{0.229} &       &       & +Geo. Cues & \textbf{0.00244} & \textbf{26.25} & \textbf{0.912} &       & \textbf{0.204} \\
\cmidrule{1-8}\cmidrule{10-16}    \multirow{3}[2]{*}{Ours} & 1     &       & 0.00394 & 24.49 & 0.886 &       & 0.644 &       & \multirow{3}[2]{*}{Input + $N(0,0.05)$} & Baseline & 0.00380 & 24.49 & 0.904 &       & 0.309 \\
          & 2     &       & 0.00319 & 25.25 & 0.903 &       & 0.420  &       &       & +Phy. Priors & 0.00247 & 26.13 & 0.909 &       & 0.232 \\
          & 3     &       & \textbf{0.00244} & \textbf{25.25} & \textbf{0.916} &       & \textbf{0.204} &       &       & +Geo. Cues & \textbf{0.00245} & \textbf{26.18} & \textbf{0.911} &       & \textbf{0.206} \\
    \bottomrule
    \end{tabular}}
          \caption{Results of models trained with different views and noisy input images. N.V represents the number of views.}
  \label{overrfitting and noisy}
\end{table*}

\subsection{Datasets and metrics}

\textbf{Datasets and evaluation metrics}.  We choose H36M~\cite{ionescu2013human3} as our primary benchmark dataset. For completeness, we also evaluate our approach on the ZJU-MoCap dataset~\cite{peng2021neural} and present a quantitative comparison to \cite{peng2021animatable} in supplementary. We benchmark our proposed HumanRecon on two tasks: generalization to novel views and poses. For image synthesis, we quantify model performance using three metrics: mean square error (MSE), peak signal-to-noise ratio (PSNR), and structural similarity index (SSIM). For geometry reconstruction, since there is no available ground truth geometry, we evaluate our approach using depth smoothness (i.e. depth error)~\cite{guo2022nerfren}.
The detailed description of benchmark datasets, implementation details and the more experimental results are presented in the supplementary.

\subsection{Ablation studies}

We conduct ablation studies to evaluate the effectiveness of each component of our proposed method on one subject ``S9" of the H36M dataset. We choose Ani-NeRF~\cite{peng2021animatable} as a default baseline model, and Ani-SDF~\cite{peng2022animatable} as an additional one.
% We first analyze the rendering capabilities of our method compared to the baseline. 
% The quantitative and qualitative results are presented in Table xxx and Figure xxx, respectively.

\textbf{The effectiveness of each component of our proposed method}. As shown in Tab.~\ref{tab:geometric and physical} and Fig.~\ref{Geometry}, we can see that both geometric constraints and physical priors can significantly improve the performance of baseline. In detail, depth cues significantly boosts reconstruction quality in terms of all evaluation metrics, especially in depth error (i.e. Dep. Err. in Tab.~\ref{tab:geometric and physical}). This shows that our approach not only obtains more faithful visual rendering results in the human reconstruction, but also better represents the underlying geometry as indicated by the lower depth error. Some improvement is also achieved when normal cues are used. With both depth and normal cues, our method achieves a better performance compared to using either one. This is also consistent with our previous analysis that global depth and local normal are complementary to each other. Different from the inherent geometric constraints above, our physical priors do not benefit the geometric results, but rather 
achieve more improvements in pixel-level image rendering. The main reason is that the additional usage of physical priors can help our method to capture finer details and alleviate overfitting on only image supervision during rendering process.
% , thus enabling the high-quality results in terms of geometry and appearance.

\textbf{The impact of perturbation on view direction}. We investigate the impact of perturbation on view direction and report the experimental results in Tab.~\ref{tab:perturbation on view}. 
We can draw the following conclusions from Tab.~\ref{tab:perturbation on view}. First, adding perturbations (noise) on the view direction can greatly improve the performance of the baseline, which is shown in the first and second blocks of Tab.~\ref{tab:perturbation on view}. Second, comparing the second and third blocks of Tab.~\ref{tab:perturbation on view}, adding noise before or after the position embedding makes no significant difference, and they are beneficial for improving the model performance.
Third, similar to Dropout~\cite{srivastava2014dropout}, the perturbation strategy can be removed during the test phase to achieve a better result.
This can be observed by comparing two modes (Train \& Test and Train) in Tab.~\ref{tab:perturbation on view}.
% is better in the test phase without adding than with adding. 
This strategy is not sensitive to hyperparameters such as noise intensity.
The last point to note is that we only change the view direction as the input to the color prediction network, not camera position ($\textbf{o}_{\text{ray}}$) and ray direction ($\textbf{d}_{\text{ray}}$), otherwise it will bring serious model degradation, as shown in the last block of the Tab.~\ref{tab:perturbation on view}.

\textbf{Comparison with traditional data augmentation methods}. 
We compare HumanRecon with traditional data augmentation methods: horizontal flips, center cropping, and adding Gaussian noise on input image. From the experimental results of the Tab.~\ref{data augmentation}, it can be seen that none of the traditional data augmentation methods can be used directly for the neural reconstruction task of dynamic human. On the contrary, our proposed method from geometric constraint and physical prior is suitable for modeling dynamic human bodies and achieves significant improvements on the original baseline.

\textbf{Overfitting problem and generalization to noisy raw data}. As shown in Tab.~\ref{overrfitting and noisy}, as our baseline, Ani-NeRF achieves the best when the number of views is 2, and faces an overfitting problem when the number of views increases to 3. However, our approach solves the overfitting problem.
The available datasets are often collected in standard scenarios with high quality. To illustrate that our method can handle more realistic scenes, we artificially generate the noisy data by adding the zero-mean noise distribution to the raw input image. As shown in Tab.~\ref{overrfitting and noisy},
our method outperforms the baseline with large margins on both noisy and noiseless data. This shows that the proposed HumanRecon is highly robust to zero-mean noise distribution and is more practical in real scenarios compared to the baseline.

\begin{table}[t]
  \centering
  \scalebox{0.75}{
    \begin{tabular}{l|cccc}
    \toprule
    \multicolumn{5}{c}{S6} \\
    Method & MSE$\downarrow$   & PSNR$\uparrow$  & SSIM$\uparrow$   & Dep. Err.$\downarrow$ \\
    \midrule
    Ani-SDF~\cite{peng2021animatable} & 0.00382 & 24.35 & 0.895  & \textbf{0.112} \\
    HumanRecon & \textbf{0.00356} & \textbf{24.61} & \textbf{0.896}  & \textbf{0.112} \\
    \midrule
    \multicolumn{5}{c}{S9} \\
    Method & MSE$\downarrow$   & PSNR$\uparrow$  & SSIM$\uparrow$  & Dep. Err.$\downarrow$ \\
    \midrule
    Ani-SDF~\cite{peng2021animatable} & 0.00264 & 26.07 & \textbf{0.918} & 0.214 \\
    HumanRecon & \textbf{0.00254} & \textbf{26.11} & 0.917  & \textbf{0.201} \\
    \bottomrule
    \end{tabular}}
 \caption{Results of models generalized to other baseline.}
  \label{tab:compatibility with others}
\end{table}

\begin{table*}[t]
  \centering
    \scalebox{0.75}{
    \begin{tabular}{lccccccccccc}
    \toprule
          &       & S1    &       &       &       & S5    &       &       &       & S6    &  \\
\cmidrule{2-4}\cmidrule{6-8}\cmidrule{10-12}          & MSE($\downarrow$)   & PSNR($\uparrow$)  & SSIM($\uparrow$)  &       & MSE($\downarrow$)   & PSNR($\uparrow$)  & SSIM($\uparrow$)  &       & MSE($\downarrow$)   & PSNR($\uparrow$)  & SSIM($\uparrow$) \\
    \midrule
    NT~\cite{thies2019deferred}    & -     & 20.98 & 0.860  &       & -     & 19.87 & 0.855 &       & -     & 20.18 & 0.816 \\
    NHR~\cite{wu2020multi}   & -     & 21.08 & 0.872 &       & -     & 20.64 & 0.872 &       & -     & 20.40  & 0.830 \\
    NARF~\cite{noguchi2021neural}  & -     & 21.41 & 0.891 &       & -     & 25.24 & 0.914 &       & -     & 21.47 & 0.871 \\
    Ani-NeRF~\cite{peng2021animatable} & -     & 22.05 & 0.888 &       & -     & 23.27 & 0.892 &       & -     & 21.13 & 0.854 \\
     \midrule
    Baseline & 0.0055 & 22.77 & 0.896 &       & 0.0046 & 23.52 & 0.893 &       & 0.0048 & 23.33 & 0.879 \\
    Baseline+Dep. & 0.0046 & 23.50  & 0.906 &       & 0.0038 & 24.23 & 0.902 &       & 0.0047 & 23.42 & 0.884 \\
    Baseline+Nor. & 0.0045 & 23.60  & 0.903 &       & 0.0043 & 23.75 & 0.896 &       & 0.0045 & 23.59 & 0.880 \\
    Baseline+Den. & 0.0047 & 23.43 & 0.901 &       & 0.0042 & 23.86 & 0.897 &       & 0.0044 & 23.71 & 0.880 \\
    Baseline+Vie. & \textbf{0.0034} & \textbf{24.72} & \textbf{0.906} &       & \textbf{0.0038} & \textbf{24.28} & \textbf{0.901} &       & \textbf{0.0037} & \textbf{24.42} & \textbf{0.883} \\
    \midrule
          &       & S7    &       &       &       & S8    &       &       &       & S9    &  \\
\cmidrule{2-4}\cmidrule{6-8}\cmidrule{10-12}     & MSE($\downarrow$)   & PSNR($\uparrow$)  & SSIM($\uparrow$)  &       & MSE($\downarrow$)   & PSNR($\uparrow$)  & SSIM($\uparrow$)  &       & MSE($\downarrow$)   & PSNR($\uparrow$)  & SSIM($\uparrow$) \\
\midrule
NT~\cite{thies2019deferred}    & -     & 20.47 & 0.856 &       & -     & 16.77 & 0.837 &       & -     & 22.96 & 0.873 \\
    NHR~\cite{wu2020multi}   & -     & 20.29 & 0.868 &       & -     & 19.13 & 0.871 &       & -     & 23.04 & 0.879 \\
    NARF~\cite{noguchi2021neural}  & -     & 21.36 & 0.899 &       & -     & 22.03 & 0.904 &       & -     & 25.11 & 0.906 \\
    Ani-NeRF~\cite{peng2021animatable} & -     & 22.5  & 0.890  &       & -     & 22.75 & 0.898 &       & -     & 24.72 & 0.908 \\
    \midrule
    Baseline & 0.0063 & 22.04 & 0.892 &       &  0.0065  &    22.81   &   0.899    &       & 0.0040 & 24.90  & 0.908 \\
    Baseline+Dep. & 0.0053 & 22.87 & 0.896 &       &  0.0054  &  23.15  & 0.902  &       & 0.0029 & 25.45 & 0.912 \\
    Baseline+Nor. & 0.0057 & 22.53 & 0.895 &       &  0.0057  &  22.96  &  0.901  &       & 0.0030 & 25.20  & 0.911 \\
    Baseline+Den. & 0.0057 & 22.61 & 0.896 &       & 0.0051 & 23.11  & 0.901  &       & 0.0032 & 25.27 & 0.912 \\
    Baseline+Vie. & \textbf{0.0038} & \textbf{24.24} & \textbf{0.899} &       & \textbf{0.0040}  &  \textbf{24.12} & \textbf{0.909}  &       & \textbf{0.0024} & \textbf{26.25} & \textbf{0.916} \\
    \bottomrule
    \end{tabular}}
  \caption{Quantitative evaluation of our method against other existing methods. Our proposed geometric cues and physical priors contribute to the performance of the baseline method
and achieve the best performance compared to other existing methods. Baseline means that  we re-train Ani-NeRF as our baseline model and report their results; ``Baseline+Dep." and ``Baseline+Nor." represent that we integrate depth and normal cues into our baseline model, respectively; ``Baseline+Den." and ``Baseline+Vie." imply that we integrate density maximization and view perturbation into our baseline model, respectively.}
  \label{tab:comparion with sota}%
\end{table*}%

\textbf{Compatibility with other baseline methods}.
Finally, to illustrate the compatibility of our approach, we choose Ani-SDF~\cite{peng2021animatable} as our baseline model. Ani-SDF models the human geometry with a signed distance field, which naturally regularizes the learned geometry and achieves the high-quality reconstruction of human bodies. Despite its power, our approach still improves rendering results (e.g. +0.25 for PSNR) on subject ``S6" and geometry quality (e.g. +0.013 for Dep. Err.) on subject ``S9", which is shown in Tab.~\ref{tab:compatibility with others}.

\begin{figure*}[t]
\begin{center}
\includegraphics[width=0.87\textwidth]{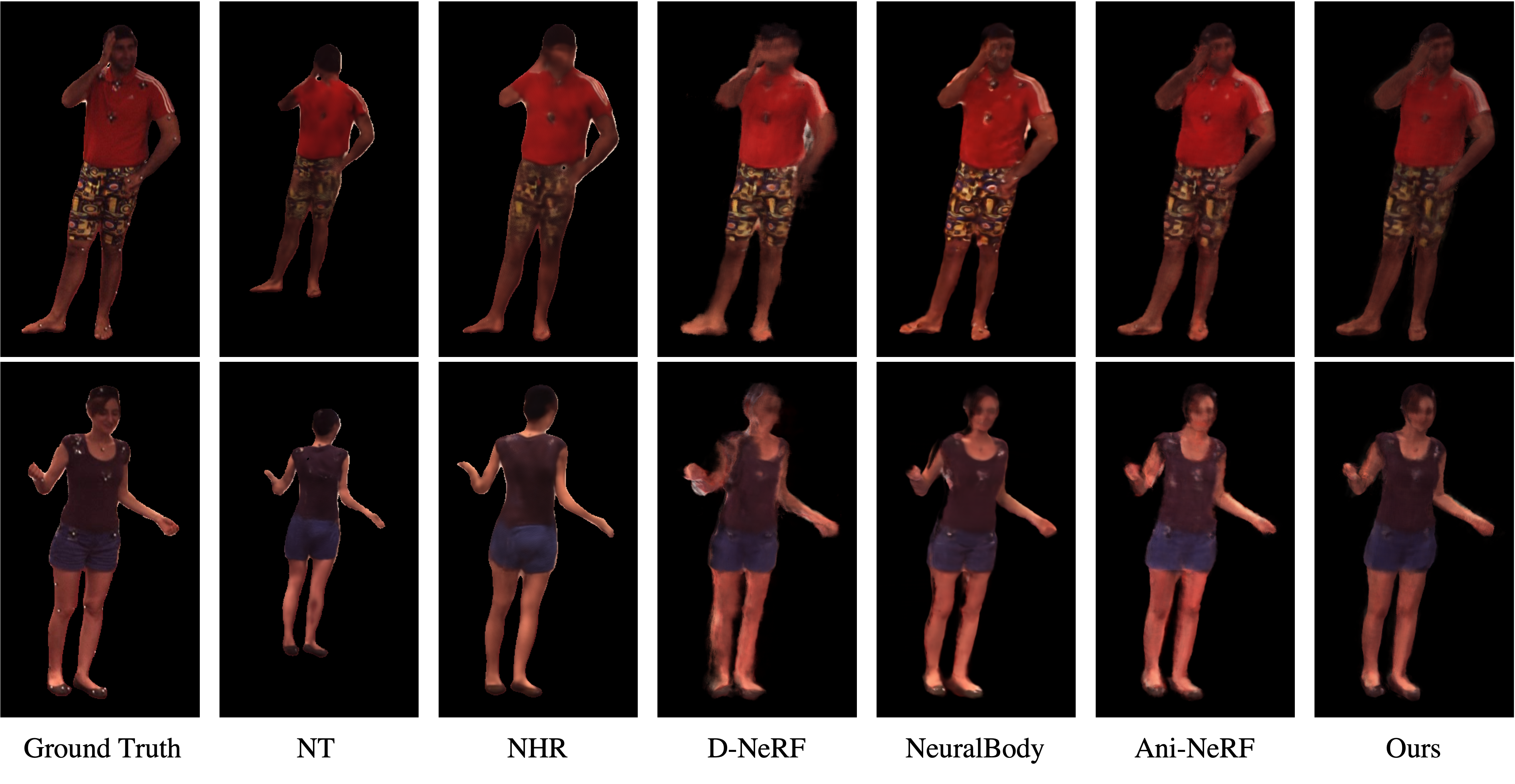}
\end{center}
\vspace{-7mm}
   \caption{Novel view synthesis results of our HumanRecon and other methods, including Ani-NeRF~\cite{peng2021animatable}, NeuralBody~\cite{peng2021neural}, D-NeRF~\cite{pumarola2021d}, NHR~\cite{wu2020multi}, and NT~\cite{thies2019deferred}.  Existing methods produce distorted rendering results and have difficult to control colors. Compared with existing methods, our method produces darker colors closer to ground truth (especially on the first row of the result) and accurately renders the target view with finer details (especially on the second row of the result). Zoom in for details.}
\label{fig:vusulize result}

\end{figure*}

\subsection{Comparison with other existing methods}

\textbf{Novel view synthesis on H36M dataset}. Tab.~\ref{tab:comparion with sota} summaries the quantitative comparison between our method with \cite{thies2019deferred}, \cite{wu2020multi}, \cite{noguchi2021neural}, \cite{peng2021animatable} on novel view synthesis of seen poses. Our proposed geometric clues and physical priors contribute to the performance of the baseline method and achieve the best performance compared to other existing methods. Compared with them,
our method is able to accurately control the viewpoint and render the human body, which is shown in Fig.~\ref{fig:vusulize result}.
This suggests that the appropriate geometry properties and physical priors are beneficial for the visual fidelity of novel views on seen poses. Due to space limitations, we present experimental results about novel view synthesis of unseen pose and another ZJU-MoCap dataset~\cite{peng2021neural} in the supplementary.

\section{Conclusion}
In this paper, we systematically explore how geometric cues and physical priors are incorporated into neural radiance fields for animatable 3D human synthesis. We show that such easy-to-obtain geometric cues provide reliable yet explicit supervision for implicit neural rendering process. As a complement, physical priors enable the network to learn view-invariant implicit representation and reduce density ambiguity along rays.  Experimental results show that the proposed approach can reconstruct accurate and smooth geometry while maintaining the high-quality rendering results on public datasets.

%%%%%%%%% REFERENCES
{\small
\bibliographystyle{ieee_fullname}
\bibliography{egbib}
}

\newpage

\noindent \textbf{Supplementary Material}

In the supplementary material, we provide the dataset description and implementation details. To show the effectiveness and generalization of our approach, we report the more qualitative and quantitative results of geometry reconstruction and image synthesis.

\section{Experiment}

\subsection{Dataset} \textbf{H36M}~\cite{ionescu2013human3} dataset captures multi-view human performers with four synchronized cameras and collects human poses with a marker-based motion capture system. We select representative actions and conduct extensive experiments on subjects ``s1", ``s5", ``s6", ``s7", ``s8", and ``s9". Among four cameras, three cameras are used for training and the remaining one is for testing. For more details settings, we refer to as~\cite{peng2021animatable}.

\textbf{ZJU-MoCap}~\cite{peng2021neural} dataset records multi-view human movements with twelve cameras and obtains human poses with the marker-based motion capture system. For the evaluation of our method on ZJU-MoCap, we choose three representative sequences datasets and conduct experiments on subjects ``313", ``315", and ``386". The four cameras are used for training and others are used for testing.

\subsection{Implementation details}
Following~\cite{peng2021animatable}, we adopt the single-level NeRF as our radiance field $F_{\sigma}$ and sample 64 points for each camera ray. Position encoding~\cite{mildenhall2021nerf} is used to encode the spatial point and the view direction. We use 6 frequencies for encoding spatial position and 4 frequencies for encoding view direction. The latent code dimension of both appearance code and blend weight field code is 128. 

\textbf{Training}. We refer to~\cite{peng2021animatable} for the training setting. Our method adopts a two-stage training pipeline. First, the parameters of the canonical human network, blend weight field, appearance code, and blend weight field code are learned over the input data. Second, under novel poses, neural blend weight fields are learned, while the remaining parameters are fixed. We use Adam optimizer~\cite{kingma2014adam} to train our model with a learning rate of $5e^{-5}$. We run all the experiments on one 2080 Ti GPU. We set the parameter of $\lambda_{\text{depth}}$ to $1$ for $L_{\text{depth}}$. For other parameters $\lambda_{\text{normal}}$ and $\lambda_{\text{normal}}$, smaller values are often more appropriate according to our empirical experiments. Note that the accuracy can be further improved when the optimal setting is used. 

\subsection{Experimental results}

\begin{table}[t]
  \centering
  \scalebox{0.9}{
    \begin{tabular}{l|cccc}
    \toprule
    \multicolumn{5}{c}{S1} \\
    Method & MSE$\downarrow$   & PSNR$\uparrow$  & SSIM$\uparrow$ & Dep. Smo.$\downarrow$ \\
    \midrule
    Baseline & 0.00706 & 21.25 & 0.867 & 0.0553 \\
    HumanRecon & \textbf{0.00552} & \textbf{22.58} & \textbf{0.876} & \textbf{0.0490} \\
    \midrule
    \multicolumn{5}{c}{S5} \\
    Method & MSE$\downarrow$   & PSNR$\uparrow$  & SSIM$\uparrow$  & Dep. Smo.$\downarrow$ \\
    \midrule
    Baseline & 0.00766 & 21.21 & 0.859  & 0.118 \\
    HumanRecon & \textbf{0.00547} & \textbf{22.69} & \textbf{0.874} & \textbf{0.110} \\
    \midrule
    \multicolumn{5}{c}{S9} \\
    Method & MSE$\downarrow$   & PSNR$\uparrow$  & SSIM$\uparrow$   & Dep. Smo.$\downarrow$ \\
    \midrule
    Baseline & 0.00445 & 23.56 & \textbf{0.878}   & 0.122 \\
    HumanRecon & \textbf{0.00383} & \textbf{24.22} & \textbf{0.878}  & \textbf{0.117} \\
    \bottomrule
    \end{tabular}}
    \caption{Results of novel pose synthesis on subjects ``S1”, ``S5”, and ``S9” of H36M dataset. Compared with the baseline, the proposed HumanRecon achieves higher image fidelity and smoother geometry results on novel poses.}
  \label{novel pose synthesis}
\end{table}%

\begin{table*}[t]
  \centering
    \begin{tabular}{crcccccccccc}
    \toprule
    \multirow{2}[3]{*}{Baseline} & \multicolumn{2}{c}{Geo. Cues} &       & \multicolumn{2}{c}{ Phy. Priors} &       & \multicolumn{3}{c}{Image} &       & \multicolumn{1}{l}{Geometry} \\
\cmidrule{2-3}\cmidrule{5-6}\cmidrule{8-10}\cmidrule{12-12}          & \multicolumn{1}{c}{Depth} & Normal &       & Vnoise & Dmax  &       & \multicolumn{1}{l}{MSE($\downarrow$)} & \multicolumn{1}{l}{PSNR($\uparrow$)} & \multicolumn{1}{l}{SSIM($\uparrow$)} &       & \multicolumn{1}{l}{Dep. Err.($\downarrow$)} \\
    $\surd$     &       &       &       &       &       &       & 0.00545 & 22.77 & 0.896 &       & 0.136 \\
    $\surd$     & \multicolumn{1}{c}{$\surd$} &       &       &       &       &       & 0.00457 & 23.50& 0.906 &       & \underline{0.120} \\
    $\surd$     &       & $\surd$     &       &       &       &       & 0.00447 & 23.60 & 0.903 &       & 0.134 \\
    $\surd$     & \multicolumn{1}{c}{$\surd$} & $\surd$     &       &       &       &       & 0.00422 & 23.83 & 0.907  &       & \underline{0.120} \\
\midrule
    $\surd$     &       &       &       & $\surd$     &       &       & 0.00344 & 24.72 & 0.906 &       & 0.135 \\
    $\surd$     &       &       &       &       & $\surd$     &       & 0.00468 & 23.43 & 0.901 &       & 0.134 \\
    $\surd$     &       &       &       & $\surd$     & $\surd$     &       & \underline{0.00341} & \underline{24.79} & \underline{0.907}  &       & 0.134 \\
    $\surd$     & $\surd$     & $\surd$    &       & $\surd$     & $\surd$     &       & \textbf{0.00334} & \textbf{24.84} & \textbf{0.909} &       & \textbf{0.119} \\
    \bottomrule
    \end{tabular}
\caption{Ablation of geometric cues and physical priors on the subject ``s1" of the H36M dataset. The proposed geometric cues and physical priors are beneficial for the reconstruction quality and rendering results of our model. ``Dmax" represents density maximization. \textbf{Bold} indicates the highest result, and \underline{underline} indicates the second highest result.}
  \label{tab:ablation_s1}
\end{table*}%

\begin{table*}[t]
  \centering
  \scalebox{0.85}{
    \begin{tabular}{clrrrrrrrrrrrr}
    \toprule
          &       &       & \multicolumn{3}{c}{313} &       & \multicolumn{3}{c}{315} &       &       & 386   &  \\
\cmidrule{1-2}\cmidrule{4-6}\cmidrule{8-10}\cmidrule{12-14}     Setting & Method &       & \multicolumn{1}{l}{MSE($\downarrow$)} & \multicolumn{1}{l}{PSNR($\uparrow$)} & \multicolumn{1}{l}{SSIM($\uparrow$)} &       & \multicolumn{1}{l}{MSE($\downarrow$)} & \multicolumn{1}{l}{PSNR($\uparrow$)} & \multicolumn{1}{l}{SSIM($\uparrow$)} &       & \multicolumn{1}{l}{MSE($\downarrow$)} & \multicolumn{1}{l}{PSNR($\uparrow$)} & \multicolumn{1}{l}{SSIM($\uparrow$)} \\
\midrule   \multirow{2}[2]{*}{Seen pose} & Ani-NeRF~\cite{peng2021animatable} &       & 0.00251 & 26.28 & 0.932 &       & 0.00687 & 21.78 & 0.848 &       & 0.00226 & 26.68 & 0.884 \\
          & HumanRecon &       & \textbf{0.00210} & \textbf{27.15} & \textbf{0.937} &       & \textbf{0.00665} & \textbf{22.01} &\textbf{ 0.853} &       & \textbf{0.00217 }& \textbf{26.85} & \textbf{0.887} \\
\cmidrule{1-6}\cmidrule{8-10}\cmidrule{12-14}    \multirow{2}[2]{*}{Unseen pose} & Ani-NeRF~\cite{peng2021animatable} &       & 0.00586 & 22.59 & 0.885 &       & 0.0171 & 17.85 & 0.790  &       & 0.00297 & 25.45 & 0.881 \\
          & HumanRecon &       & \textbf{0.00564} & \textbf{22.83} & \textbf{0.887} &       & \textbf{0.0156} & \textbf{18.29} &\textbf{ 0.800}   &       & \textbf{0.00296} & \textbf{25.52 }& \textbf{0.882} \\
    \bottomrule
    \end{tabular}}
  \caption{Quantitative results of models on ZJU-MoCap seen poses and unseen poses. Compared with Ani-NeRF~\cite{peng2021animatable}, our approach achieves a better performance of novel view synthesis on seen and unseen poses.}
  \label{tab:zju}
\end{table*}

\begin{figure*}[t]
\begin{center}
\includegraphics[width=1.0\textwidth]{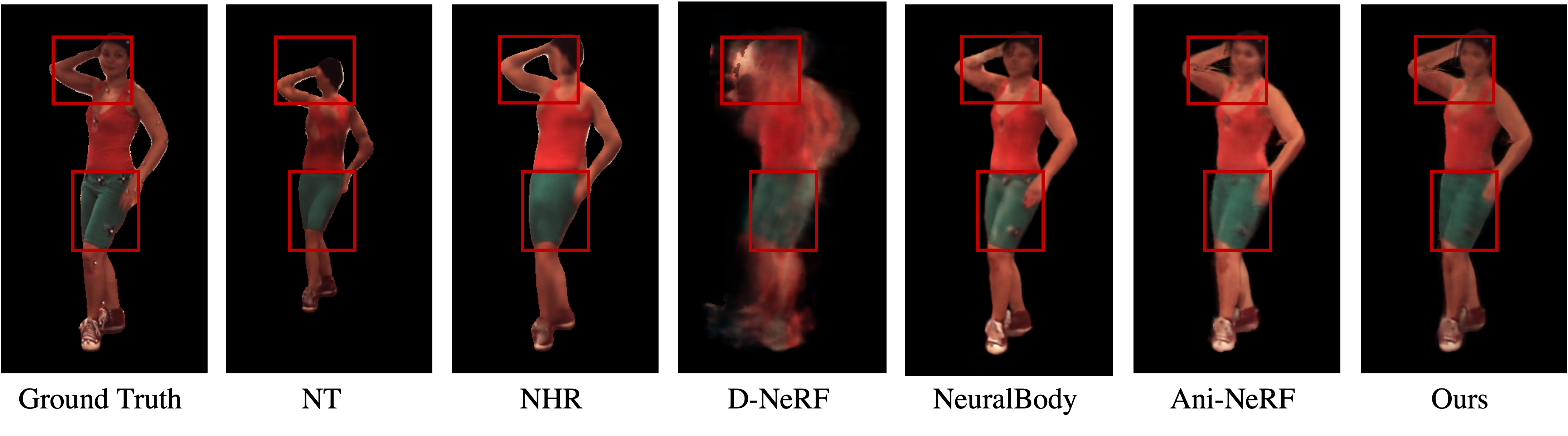}
\end{center}
\vspace{-7mm}
   \caption{Novel view synthesis results of our HumanRecon and other methods on subject ``s7" of H36M dataset, including Ani-NeRF~\cite{peng2021animatable}, NeuralBody~\cite{peng2021neural}, D-NeRF~\cite{pumarola2021d}, NHR~\cite{wu2020multi}, and NT~\cite{thies2019deferred}. Compared with existing method, our approach achieves higher visual quality with fewer artifacts while preserving the details on the faces and arms of the subjects. Zoom in for details.}
\label{visualize_supp}
\end{figure*}

\begin{figure*}[t]
\begin{center}
\includegraphics[width=1.0\textwidth]{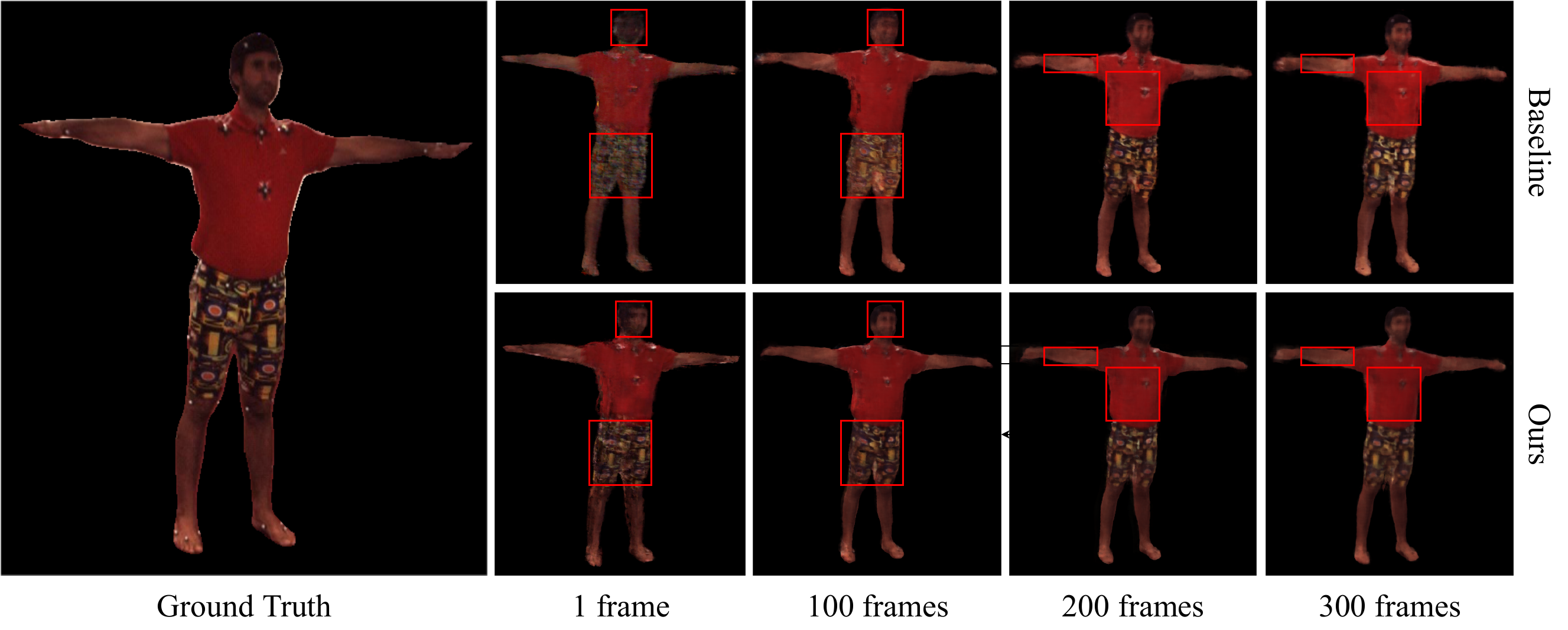}
\end{center}
\vspace{-7mm}
   \caption{Comparison between models trained with different numbers of video frames on the subject “S9” of H36M dataset.}
\label{trainnum}
\end{figure*}

\begin{figure*}[t]
\begin{center}
\includegraphics[width=1.0\textwidth]{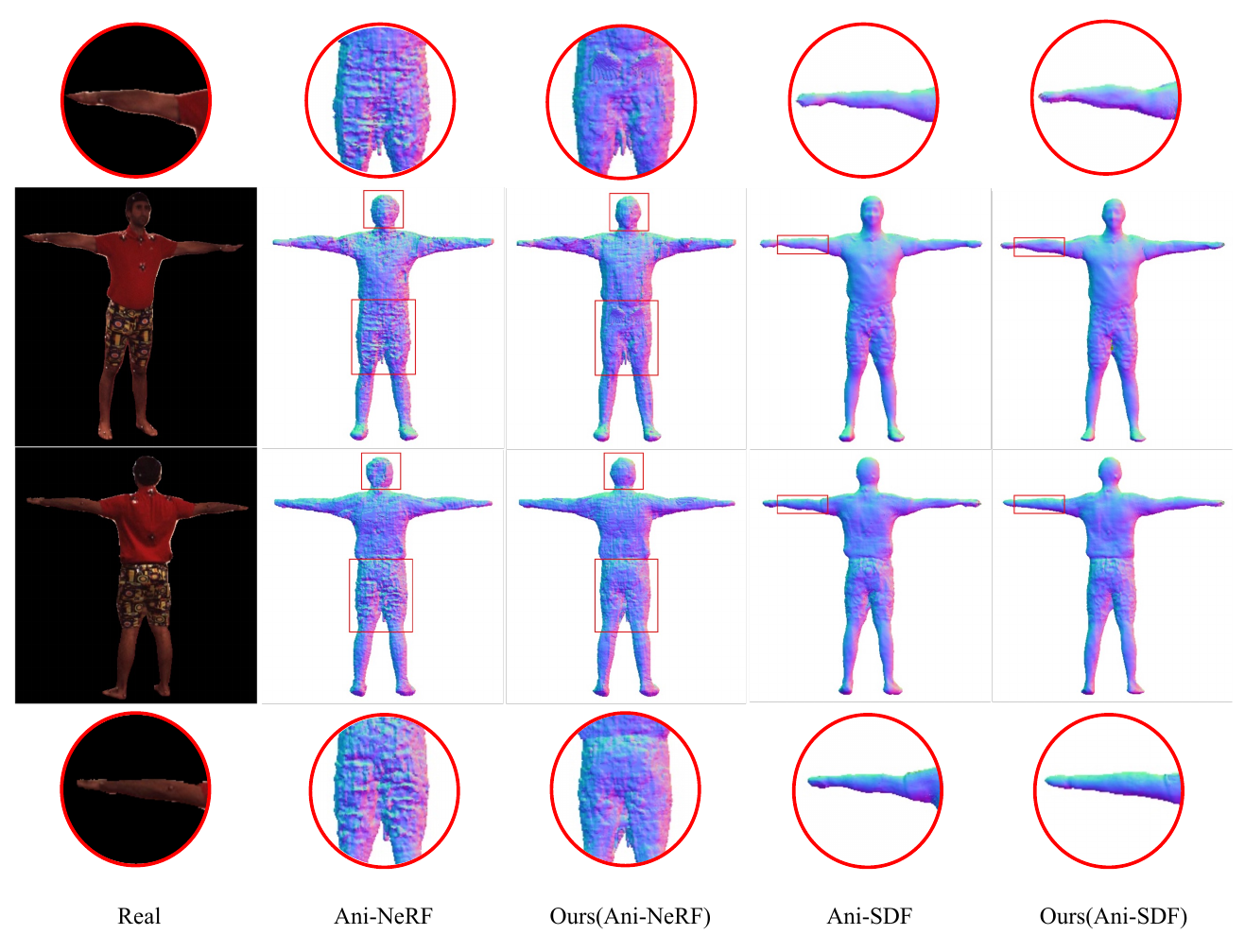}
\end{center}
\vspace{-7mm}
   \caption{Geometry reconstruction results on different baseline models.  compared with Ani-NeRF~\cite{peng2021animatable} and Ani-SDF~\cite{peng2022animatable}, our approach not only reconstructs smoother geometry results with high-quality but also captures more geometry details (e.g. arm details). Zoom in for details.}
\label{geometry2}
\end{figure*}

\textbf{Novel view synthesis on H36M dataset}. Novel view synthesis results on seen pose have been discussed in the main paper. Next, we investigate our proposed HumanRecon for novel view synthesis on unseen poses. Tab.~\ref{novel pose synthesis} presents a quantitative comparison to the baseline. Compared with the baseline, the proposed HumanRecon achieves higher image fidelity and better consistency across novel views. This indicates that our approach can be generalized to unseen poses and suited for representing animatable human model. As shown in Fig.~\ref{visualize_supp}, our method produces higher visual quality with fewer artifacts than existing methods~\cite{peng2021animatable,peng2021neural,pumarola2021d,wu2020multi,thies2019deferred}, which also indicates a better correspondence across frames.

\textbf{Novel view synthesis on the ZJU-MoCap dataset}. We present additional quantitative results on ZJU-MoCap dataset. Compared with the baseline model, our approach obtains a better performance on seen and unseen poses in terms of MSE, PSNR, and SSIM, which is shown in Tab.~\ref{tab:zju}. For example, the proposed HumanRecon achieves 0.87\% improvement for PSNR on subject ``313" with seen poses than Ani-NeRF~\cite{peng2021animatable}. We argue this performance benefits from the additional prior knowledge of geometry and physics in our method while enhancing the learning for neural implicit representation.

\textbf{Ablation on subject ``s1" of the H36M dataset}. We further conduct some ablation studies to verify the effectiveness of each component of our method on subject ``s1" of the H36M dataset. Experimental results are shown in Tab.~\ref{tab:ablation_s1}.  From the results of Tab.~\ref{tab:ablation_s1}, we can observe that geometric cues achieve 1.06\% improvement for PSNR and 0.16 improvement for Dep. Err. than the baseline. With this module, our method has explicit geometry supervision and reduces the geometrically ambiguity, therefore enhancing the rendering capability of the human reconstruction model. Similar performance can be found in the physical priors module. With both modules, our method reaches the best performance by successfully constraining both geometry and appearance.

\textbf{Comparison between models trained with different numbers of video frames on the subject “S9”.} In this subsection, we present the comparison between models trained with different numbers of video frames. For this setup, we do not modify our method and all hyperparameters are the same as before. 
For comparison, we train the baseline and our model with 1,
100, 200, and 300 video frames and test the models on the same pose sequence. We present the qualitative results in Fig.~\ref{trainnum}. Due to both geometrical and physical constraints, our method better capture clothing color, arms, and face details which the baseline suffers from. The stable performance is achieved for our model at 100 video frames, while the baseline model achieves stable performance with 200 video frames.

\textbf{Geometry reconstruction results on different baseline models}. We adopt Ani-NeRF~\cite{peng2021animatable} and Ani-SDF~\cite{peng2022animatable} as our baseline model and present the reconstruction results of our method in comparison with the two baseline models. As shown in Fig.~\ref{geometry2}, the overall results of Ani-SDF based methods are better than that of Ani-NeRF based method. Specifically, compared with Ani-NeRF~\cite{peng2021animatable} and Ani-SDF~\cite{peng2022animatable}, our approach not only reconstructs smoother geometry results with high-quality but also captures more geometry details (e.g. arm details).

\end{document}